\documentclass[sigconf]{acmart}

\makeatletter
\def\input@path{{acm/}{./}}
\makeatother
\usepackage{xspace}
\usepackage{siunitx}
\DeclareSIUnit{\fps}{fps}
\usepackage[inline]{enumitem}
\usepackage{multirow}

\newcommand{\ie}{\emph{i.e.}, }
\newcommand{\eg}{\emph{e.g.}, }

\newcommand{\vxv}[2]{$#1\!\times\!#2$}
\AtBeginDocument{%
  }

\copyrightyear{2025}
\acmYear{2025}
\makeatletter
\def\@ACM@copyright@check@cc{}
\makeatother
\setcopyright{cc}
\setcctype{by}
\acmConference[MM '25]{Proceedings of the 33rd ACM International Conference on Multimedia}{October 27--31, 2025}{Dublin, Ireland}
\acmBooktitle{Proceedings of the 33rd ACM International Conference on Multimedia (MM '25), October 27--31, 2025, Dublin, Ireland}
\acmDOI{10.1145/3746027.3758153}
\acmISBN{979-8-4007-2035-2/2025/10}

\begin{document}

\title[GenStream]{GenStream: Semantic Streaming Framework for Generative Reconstruction of Human-centric Media}

\author{Emanuele Artioli}
\email{emanuele.artioli@aau.at}
\orcid{0009-0007-9921-0006}
\affiliation{
  \institution{Alpen-Adria-Universitaet}
  \city{Klagenfurt}
  \state{Kaernten}
  \country{Austria}
}
\author{Daniele Lorenzi}
\email{daniele.lorenzi@aau.at}
\orcid{0000-0003-3689-6315}
\affiliation{
  \institution{Alpen-Adria-Universitaet}
  \city{Klagenfurt}
  \state{Kaernten}
  \country{Austria}
}
\author{Shivi Vats}
\email{shivi.vats@aau.at}
\orcid{0009-0003-4374-9329}
\affiliation{
  \institution{Alpen-Adria-Universitaet}
  \city{Klagenfurt}
  \state{Kaernten}
  \country{Austria}
}
\author{Farzad Tashtarian}
\email{farzad.tashtarian@aau.at}
\orcid{0000-0002-5584-6690}
\affiliation{
  \institution{Alpen-Adria-Universitaet}
  \city{Klagenfurt}
  \state{Kaernten}
  \country{Austria}
}
\author{Christian Timmerer}
\email{christian.timmerer@aau.at}
\orcid{0000-0002-0031-5243}
\affiliation{
  \institution{Alpen-Adria-Universitaet}
  \city{Klagenfurt}
  \state{Kaernten}
  \country{Austria}
}

\begin{abstract}
Video streaming dominates global internet traffic, yet conventional pipelines remain inefficient for structured, human-centric content such as sports, performance, or interactive media. Standard codecs re-encode entire frames, foreground and background alike, treating all pixels uniformly and ignoring the semantic structure of the scene. This leads to significant bandwidth waste, particularly in scenarios where backgrounds are static and motion is constrained to a few salient actors. We introduce GenStream, a semantic streaming framework that replaces dense video frames with compact, structured metadata. Instead of transmitting pixels, GenStream encodes each scene as a combination of skeletal keypoints, camera viewpoint parameters, and a static 3D background model. These elements are transmitted to the client, where a generative model reconstructs photorealistic human figures and composites them into the 3D scene from the original viewpoint. This paradigm enables extreme compression, achieving over 99.9\% bandwidth reduction compared to HEVC for the continuous data stream. We partially validate GenStream on Olympic figure skating footage and demonstrate potential for high perceptual fidelity under minimal data. While acknowledging the significant computational costs shifted to the client and challenges in generalization, GenStream opens new directions in volumetric avatar synthesis, canonical 3D actor fusion across views, and personalized viewing experiences, laying the groundwork for scalable, intelligent streaming in the post-codec era.
\end{abstract}

\renewcommand{\shortauthors}{Emanuele Artioli, Daniele Lorenzi, Shivi Vats, Farzad Tashtarian, \& Christian Timmerer}

\ccsdesc[500]{Computing methodologies~Object identification}
\ccsdesc[500]{Computing methodologies~Artificial intelligence}
\ccsdesc[300]{Information systems~Online analytical processing}
\ccsdesc[500]{Information systems~Multimedia streaming}
\ccsdesc[300]{Computing methodologies~Concurrent algorithms}
\ccsdesc[500]{Computing methodologies~Image compression}

\keywords{HTTP adaptive streaming, Generative AI, End-to-end architecture, Quality of Experience}


\maketitle

\section{Introduction}

Video streaming plays a pivotal role in digital media consumption and accounts for the majority of global Internet data traffic~\cite{sandvine}. Specifically, major live international events such as the Olympic Games attract millions of viewers worldwide, generating massive volumes of data as high-definition video streams are delivered simultaneously to heterogeneous devices~\cite{olympics2024}. With the growing appetite for immersive and interactive content, the limitations of existing video streaming paradigms are becoming increasingly evident.

HTTP Adaptive Streaming (HAS) has become the de facto standard for video delivery over the Internet, dynamically adjusting video quality based on real-time network conditions~\cite{8424813, nguyen2022dofp+}. In HAS, video content is partitioned into segments of fixed duration (\eg 2 to 10 seconds~\cite{zhang2019towards}), each encoded independently using a video codec such as Advanced Video Coding (AVC)~\cite{h264_overview} or its successor, High Efficiency Video Coding (HEVC)~\cite{h265_overview}. To support adaptive switching, each segment is encoded into several quality representations, where each representation is defined by a specific bitrate and resolution pair, collectively referred to as the bitrate ladder.
While modern codecs are effective at reducing redundancy through intra-frame (spatial) and inter-frame (temporal) compression~\cite{h264_overview, h265_overview}, they operate under strict constraints: each segment is independently encoded and, hence, can only reference frames within its temporal window, limiting inter-frame prediction to short-term dependencies~\cite{zhang2020interframe}. This means that when long-term repetitive patterns occur, the codec repeatedly re-encodes similar content, failing to capitalize on the inherent structural redundancy~\cite{skupin2021open}.
A compelling example of such inefficiency can be observed in sporting events, where athletes move against a generally static or slowly evolving background, such as a stadium, lighting fixtures, and surrounding audience. Despite the structural consistency of such scenes, conventional codecs repeatedly re-encode the same portions of background every time the camera passes over them. This redundancy issue is further exacerbated in multi-camera productions, where identical scenes captured from different viewpoints are encoded independently, squandering opportunities to exploit cross-view redundancy~\cite{zhang2018adaptive, zhu2023implicitexplicitintegratedrepresentationsmultiview}.
To address these inefficiencies, we propose GenStream, a framework designed for live streaming that redefines how such events can be streamed and experienced. Instead of transmitting full video frames, GenStream captures skeleton-based keypoints (\eg body landmarks) of the performers with their relative bounding box and sends this compact representation to the client. 
The server then tackles a critical challenge: estimating the original camera viewpoint for each frame, which is essential for replicating the broadcast viewpoint. This pose, along with the performer keypoints, is sent to the client. At the client side, a second core challenge arises: generating a photorealistic rendering of the performer from sparse input using a generative model such as a conditional Generative Adversatial Network (cGAN)~\cite{kurupathi2023cgan} or diffusion model~\cite{podell2023sdxlstablediffusion}. The synthesized figure is then composited into a pre-scanned 3D model of the venue, and rendered from the provided viewpoint to faithfully recreate the original scene.
This paradigm shift offers dramatically reduced bandwidth usage, and support for personalized viewing experiences, such as a stylistic customization of the performers or points of view that differ from what the real cameras are capturing. Such personalization is a critical factor in media delivery, as understanding and catering to individual user sensitivities has been shown to directly impact and improve user engagement~\cite{digitwise}.

According to our preliminary results, GenStream can reduce the required live streaming bitrate by over 99.9\% compared to a baseline HEVC-encoded video, assuming the 3D representation of the venue and generative model weights are pre-distributed to clients. This impressive reduction in bandwidth enables a stall-free streaming experience at the massive scale required for worldwide sporting events, even under constrained or fluctuating network conditions. At the same time, GenStream preserves high visual fidelity through client-side synthesis, albeit at the expense of increased computational complexity.

\section{Related Work}
Our work is positioned at the intersection of traditional video streaming, semantic video analysis, and generative synthesis. We build upon decades of work in video compression while taking inspiration from emerging paradigms in volumetric and model-based media transmission.

\subsection{Traditional Video Streaming and Codecs}
Video streaming today is predominantly driven by standard protocols such as HTTP Adaptive Streaming (HAS)~\cite{Stockhammer2011dash}, which underlie systems like HLS~\cite{AppleHLS} and MPEG-DASH~\cite{dashjs}. These methods divide content into short segments and offer multiple quality tiers to allow the client to adapt playback to current network conditions. Combined with client-side adaptive bitrate (ABR) logic, these systems ensure playback continuity and have been highly successful in scaling video delivery across heterogeneous networks and devices.
Compression in these pipelines is handled by increasingly sophisticated codecs, such as AV1~\cite{av1_overview} and VVC (H.266)~\cite{vvc_overview}, which improve upon earlier standards in terms of compression efficiency, often achieving up to 50\% savings. These advances are achieved through better motion prediction, transform coding, and entropy models. However, such gains come at a significant computational cost, especially on the encoder side, with high-resolution or real-time use cases facing increased energy and latency burdens~\cite{av1_overview, vvc_overview}.
Despite decades of progress, traditional codecs treat all pixels equally and operate at the block level, agnostic to scene semantics. As a result, even small improvements in compression now require disproportionate increases in complexity, yielding diminishing returns for practical deployment.

\subsection{Emerging Architectures for Video Encoding}
To address some of the inefficiencies of standard codecs, recent research has explored novel encoding paradigms that move beyond traditional block-based compression. One approach augments existing codecs with machine learning components to improve rate control, motion estimation, or mode decisions~\cite{liu2020deep, lu2019dvc}. These techniques often integrate seamlessly with legacy pipelines, but remain bound by the structure of the original codec.
A more radical line of work proposes fully neural video compression systems, in which residuals and motion fields are learned end-to-end by deep networks~\cite{lu2019dvc, chen2021nerv}. While promising in terms of visual quality, these models are often too computationally intensive for real-time use or deployment on resource-limited clients.
Similar techniques have been developed for the 3D media landscape, i.e., volumetric video streaming, which aims to enable immersive, free-viewpoint experiences. Research in this area has explored transmission schemes for point cloud videos, using neural networks to extract and reconstruct features to reduce bandwidth~\cite{semantic2022volumetric}, and allow for photorealistic rendering from arbitrary viewpoints~\cite{ai2024volumetric}.
More recently, data-driven representations like Neural Radiance Fields (NeRF)~\cite{mildenhall2021nerf} and 3D Gaussian Splatting (3DGS)~\cite{zhou20243dgs} have opened new possibilities for compact and continuous scene representations. While typically used for static scene modeling, they provide new pathways for transmitting visual content as a structured function of space and viewpoint.
Complementary to these efforts, other works focus on client-side enhancement techniques that improve the subjective quality of the received video. Methods such as frame interpolation~\cite{niklaus2020softmax, xu2019quadratic} and super-resolution~\cite{zhu2020semanticsr, ledig2017superresolution, zhang2017fastsuperresolution} can increase perceived temporal or spatial fidelity without increasing the transmitted bitrate.
However, these emerging techniques all share a common trait: they treat video as a dense signal composed of low-level features such as pixels, motion vectors, or residuals. They do not leverage higher-level understanding of scene composition, such as objects, or actions, and thus miss a key opportunity for more structured and efficient encoding.

\subsection{Object-Centric Representations and Semantic Streaming}
The core idea of transmitting semantic information instead of raw pixels has been explored extensively, particularly in the context of video conferencing and volumetric video.
Early work in object-based coding, such as MPEG-4 Part 2~\cite{pereira2002mpeg}, allowed for the separate encoding of video object planes (VOPs), but saw limited adoption due to the complexity of robust, real-time segmentation. 
More recently, the rise of deep learning has revitalized semantic approaches. For instance, ELVIS~\cite{elvis} uses semantic segmentation to identify and remove low-priority background regions before transmission, using generative inpainting on the client to reconstruct them. This, however, still relies on traditional codecs for both foreground and higher-priority background regions.

Semantic video streaming shifts the encoding focus from raw pixels to meaningful scene components~\cite{zhang2020semantic3d}. This perspective recognizes that not all regions of a video are equally important: agents (such as humans, vehicles, or other active foreground elements) are typically of greater perceptual and narrative significance than background elements. By decoupling foreground and background and encoding them differently, streaming systems can better match perceptual salience to bitrate allocation. 
Systems like SemConf~\cite{liu2023semconf} transmit facial landmarks and background masks to drive a generative model on the client side, achieving significant bandwidth savings for talking-head scenarios. Similarly,~\cite{talking2021head} proposed a one-shot free-view synthesis method for video conferencing, while~\cite{wireless2022semantic} explored wireless semantic communications for the same application. These systems demonstrate the viability of generative reconstruction from sparse metadata.
To enable semantic encoding, a variety of tools have emerged. Object detection and tracking algorithms allow for persistent identification of foreground elements across time~\cite{meinhardt2022trackformer, zhang2022bytetrack}. Instance segmentation networks can further delineate objects from their surroundings, enabling region-specific treatment during encoding~\cite{redmon2016yolo}. These components form the foundation for object-centric video understanding.
While these techniques have seen some integration in video conferencing systems, where avatar-based transmission and background substitution are common, their use remains tightly scoped to face-to-face communication scenarios. In broader video streaming, particularly for entertainment, sports, or interactive media, semantic approaches are largely absent. The vast majority of streaming systems continue to transmit raw pixels rather than structured representations, leaving the potential of object-level encoding underutilized.

\subsection{Generative Reconstruction}
Generative models have recently demonstrated remarkable capabilities in synthesizing photorealistic video content from abstract inputs such as pose skeletons, semantic maps, or depth~\cite{bazarevsky2020blazepose, cao2017realtime}. This progress has been fueled by advances in both image and video generation, ranging from GAN-based architectures to diffusion models, and has enabled applications in human reanimation, style transfer, and controllable video generation~\cite{ledig2017superresolution, kurupathi2023cgan, ho2020diffusion, podell2023sdxlstablediffusion}.
In the context of streaming, these models offer an opportunity to decouple the representation of motion and appearance. Instead of transmitting dense pixel data, a system may instead send a sparse representation, such as 2D keypoints or depth maps, and rely on a generative model at the client to reconstruct the full image content. This approach opens the door to radically more efficient pipelines.

To overcome current limitations and develop a truly semantic streaming system, future efforts can draw from advances in camera motion estimation and 3D scene reconstruction. Accurate tracking of camera viewpoint~\cite{mur2015orbslam, davison2007monoslam, vijayanarasimhan2017sfm} can enable dynamic viewpoint control on the client. Meanwhile, neural scene representations allow the static background to be precomputed and encoded efficiently as a view-consistent 3D asset. When combined with semantic foreground representations, these modalities enable a modular and compositional approach to streaming where agents, environments, and motion can each be transmitted and reconstructed independently.

\section{Problem Description and Motivation}

\begin{figure*}
    \centering
    \includegraphics[width=1\linewidth]{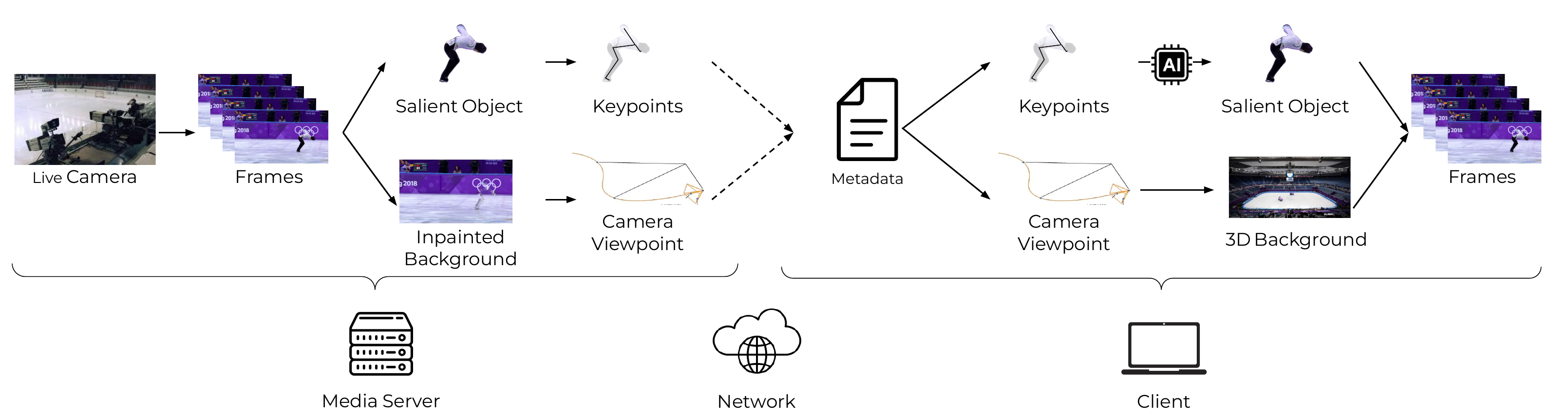}
    \caption{Conceptual architecture of GenStream.}
    \Description{Conceptual architecture of GenStream.}
    \label{fig:overview}
\end{figure*}

To evaluate the effectiveness of our approach in terms of video quality and bandwidth reduction, we first establish a baseline by measuring the performance of current state-of-the-art codecs under constrained bandwidth conditions. We apply this comparison to a representative 1080p video scene from an artistic ice skating competition, recorded at a frame rate of \SI{30}{\fps} and \SI{50}{\second} in duration.
As our reference codec, we select the High-Efficiency Video Codec (HEVC)~\cite{h265_overview} due to its widespread adoption~\cite{bitmovin-report} and high compression efficiency~\cite{h265_overview}.
The target video~\cite{olympics2018video} is encoded at a resolution of \vxv{1920}{1080} with a bitrate of \SI{5.8}{Mbps}.
Artistic ice skating competitions present a compelling opportunity for compression via semantic and structural decomposition. A typical broadcast sequence features:
\begin{enumerate*}[label=\textit{(\roman*)}]
\item a static or slowly evolving background (the ice rink, boards, and stadium interior), and
\item one or two foreground objects of interest (an individual skater or a couple of skaters).
\end{enumerate*}
Rather than encoding full video frames pixel-by-pixel, the scene can be decomposed into a persistent 3D background and a set of dynamic pose-driven elements. Therefore, assuming we possess the 3D point cloud of the stadium, to reconstruct the scene, we must first estimate the camera viewpoint \( T \) from the input video scene~\cite{hartley2004multiview}:

\[
\mathbf{T} =
\begin{bmatrix}
\mathbf{R} & \mathbf{t} \\
\mathbf{0}^{\top} & 1
\end{bmatrix}
\in \mathbb{R}^{4 \times 4}
\]

Here, \( \mathbf{R} \in  \mathbb{R}^{3 \times 3}\) is the rotation matrix, and \( \mathbf{t} \in \mathbb{R}^{3}\) is the camera center in world coordinates. The bottom row \( [\mathbf{0}^{\top} 1] \) makes it a \vxv{4}{4} homogeneous transformation matrix. 
This 3D scene needs to be rendered at the client side within the Unity~\cite{unity} framework. Since Unity represents any rotation via a quaternion \( \mathbf{q} \in \mathbb{R}^4 \), which represents a mathematically convenient alternative to the euler angle representation, we can convert  \(\mathbf{R}\) to  \( \mathbf{q} \) using quaternion algebra~\cite{voight2021quaternion}. This way, we reduce the required values to represent \( \mathbf{T} \) from \num{12} (\num{3} for \( \mathbf{t} \) and \num{9} for \( \mathbf{R} \)) to \num{7} (\num{3} for \( \mathbf{t} \) and \num{4} for \( \mathbf{q} \)). With the camera viewpoint \( \mathbf{T} \) and the 3D point cloud of the stadium, the background can be easily reconstructed.

The foreground, on the other hand, includes dynamic objects, \ie the skaters, that can be cropped out of the captured 2D video, delivered as separate streams to the clients, and overlaid in the camera view based on their original position. Although it requires less bandwidth, this approach presents several technical challenges. Since the bounding box used to extract each skater varies over time, the resulting cropped figures would have non-uniform spatial resolutions, complicating real-time rendering and synchronization on the client side. Padding cropped frames with black pixels to a fixed resolution is bandwidth-inefficient, similar to codec overhead from aligning with fixed-size coding blocks~\cite{li2010rate}. Transmitting each crop as an independent image sequence is even worse, as it forgoes inter-frame compression gains~\cite{salih2024impact}. These inefficiencies led us to replace pixel-based representations with a minimal structured alternative.
Rather than transmitting full pixel-based representations of the skaters, we transmit a compact set of skeletal keypoints \(\mathcal{S} = \{(x;y)\ |~x \leq W; y \leq H\}\), where \( W \) and \( H \) denote the width and height of the video frame, respectively. These keypoints efficiently encode the motion and structural pose of the skaters over time. The total number \( V \) of values necessary to reconstruct the scene is:

\begin{equation}
    V = (2\times|\mathcal{S}|+|\mathcal{B}|+|\mathbf{T}|)\times{F}\times{L}
\end{equation}

where \( \mathcal{B} \) is the per-frame set of values representing the bounding box of the skater, \( |\mathbf{T}| \) is the minimum number of parameters to represent the camera viewpoint \( \mathbf{T} \), \( F \) is the frame rate of the video in frames per second, and \( L \) is the length of the video in seconds.
It is worth noting that the generative model can be transmitted to the client before the stream takes place and, therefore, does not contribute to the streaming bandwidth.

Depending on the resolution of the input video, values in \( \mathcal{B} \) and \(\mathcal{S}\) can be efficiently represented using \SI{11}{\bit} (\vxv{1920}{1080}) or \SI{12}{\bit} (\vxv{3840}{2160}) integers. For the camera viewpoint \( T \), we consider \SI{32}{\bit} floating-point values.
The motion of each ice skater throughout the whole video, encoded as 17 keypoints~\cite{cao2017realtime}, requires \SI{64}{\kilo\byte} (plus less than \SI{4}{\kilo\byte} for the bounding boxes), while camera viewpoints require additional \SI{42}{\kilo\byte}. The total data transmitted is, therefore, approximately \SI{110}{\kilo\byte}, corresponding to a 99.9\% reduction in bandwidth compared to the original \SI{134}{\mega\byte} HEVC-encoded video. It is important to note that this calculation pertains only to the continuous data stream and does not include the one-time, pre-stream distribution of the 3D venue model and generative model weights, which constitute a significant initial payload.
Despite this impressive compression, perceptual coherence is preserved. The reconstructed stadium maintains spatial realism, while the skaters’ poses and trajectories are accurately rendered in appearance and timing. Although conventional quality metrics like VMAF may decline due to the lack of pixel-level fidelity, the semantic content and visual plausibility remain intact, especially for downstream use cases such as highlight replays, performance analysis, or live streaming in low-bandwidth environments. A formal perceptual study is required to quantify this trade-off, which remains a key area for future work.

\section{System Design}

GenStream proposes a radical departure from conventional video pipelines by eliminating pixel-level redundancy at the server and instead transmitting only a sparse, generative codebook of the scene. The system rethinks the encoding of human-centric motion and background information as two disjoint but complementary generative tasks: appearance reconstruction and spatiotemporal placement.
Unlike conventional codecs or even object-aware compression pipelines, GenStream treats video as a generative instruction set rather than a sequence of images. At its core is a hybrid pipeline that combines object understanding, 3D scene modeling, and generative synthesis.
Figure~\ref{fig:overview} illustrates the architecture of GenStream, broken down into its server-side and client-side pipelines.

\subsection{Server-side GenStream}
The server-side component transforms raw videos into a set of semantically meaningful layers: dynamic object keypoints, background images, and camera points of view.
We describe four key stages below:

\subsubsection*{Salient Actor Extraction}
The pipeline begins by identifying and segmenting salient actors from each frame. This is accomplished via Artificial Intelligence (AI) through object detection and instance segmentation models.
To lower the chance of misclassifying a spectator for the actor, we segment only the largest detected human in each frame. This produces an alpha mask that is used to construct two complementary Red Green Blue Alpha (RGBA) layers per frame:
\begin{enumerate*}[label=\textit{(\roman*)}]
    \item A foreground image containing only the actor, and 
    \item a background image with the actor masked out.
\end{enumerate*}

The segmentation step acts as an attention filter that reduces each frame to its most perceptually critical content, discarding clutter, crowds, and occlusions.

\subsubsection*{Keypoint-based Pose Encoding}
For each segmented actor, GenStream extracts high-resolution keypoint trajectories across time. These form the structured motion encoding that guides generative synthesis on the client side.
Our system uses state-of-the-art pose estimation models to track anatomical keypoints of the actor, and encodes them into a temporal stream. The keypoints serve as lightweight motion blueprints for actor synthesis downstream.
Unlike traditional video codecs that preserve every pixel, GenStream transmits only these skeletal pose traces, implicitly entrusting the client to fill in visual details through generative inference.

\subsubsection*{Background Inpainting and 3D Reconstruction}
The server assembles a clean, actor-free background model from the masked RGBA background layers. Regions previously occluded by actors are inpainted using context-aware algorithms to restore full-frame realism and aid the Structure-from-Motion (SfM) algorithm to infer camera viewpoint. These poses allow the client to re-render dynamic actors within a coherent 3D scene.
This use of background-only frames to drive global camera understanding is a key innovation of GenStream: it decouples the spatiotemporal understanding of scene layout from the actors themselves, enabling more robust reconstructions.
It is important to note that this step becomes essential when the 3D model of the venue and camera viewpoints are not provided by the competition organizers apriori.

\subsubsection*{Encoding for Transmission}
Finally, the server packages actor keypoints and camera viewpoints coordinates into a single metadata file. The structured nature of these coordinates allows for a highly optimized compression logic, resulting in massive bitrate gains when compared to the traditional block-based video encoding.

\subsection{Client-side GenStream}
The client-side of GenStream is responsible for converting the structured transmission into a visually plausible, temporally coherent video. Instead of decoding pixels, it interprets abstract control representations (keypoints + poses) and re-renders the video using a generative prior.
Three key modules are involved:

\subsubsection*{Skeleton-Driven Actor Reconstruction}
The received keypoint stream is decoded into a full-body actor reconstruction via a conditional generative model. 
This step synthesizes full-frame RGB actors at arbitrary resolutions and quality levels, depending on the client's capabilities and time constraints.

\subsubsection*{Viewpoint-Aware Scene Recomposition}
Using the 3D camera viewpoints and original spatial metadata, the client reinserts each synthesized actor into the 3D background, which can be a custom-made representation sent to the client prior to the streaming event, or extracted by the same algorithm that inferred the camera viewpoints. The composition is performed frame-by-frame, guided by the temporal evolution of the keypoints and viewpoint metadata.

\subsubsection*{Frame Synthesis and Playback}
The final video is reconstructed by compositing synthesized actors into the 3D background. This reconstructed sequence is passed to the video player for smooth, low-latency playback.
By relying on learned priors and sparse representations, GenStream is robust to bandwidth drops, jitter, or loss of intermediate frames. It behaves more like a generative animation system than a classic video streaming system.

\subsection{Design Summary}
In summary, GenStream replaces blocks of pixels with poses, shifting the core payload from image data to motion semantics, separates actor and background processing, enabling independent treatment and optimization of each layer, and uses client-side generative models to recreate known figures, dramatically reducing the required bandwidth.
This design explores an alternative to the traditional pixel-based video stack, leveraging semantic understanding to create a blueprint for generative streaming systems. However, this approach introduces a fundamental trade-off: it drastically reduces network bandwidth at the cost of significantly increased computational complexity on the client device. Our proof-of-concept implementation relies on a powerful GPU for client-side rendering, a configuration that is not representative of typical consumer devices like smartphones or smart TVs. GenStream's client-side burden is a major limitation of the current framework. Future work must explore strategies to mitigate this cost, such as generative model quantization, knowledge distillation to create lightweight client models, or offloading synthesis to edge servers. Without such optimizations, the practical applicability of GenStream remains limited to powerful client devices.

\section{Implementation}

\begin{figure*}
    \centering
    \includegraphics[width=1\linewidth]{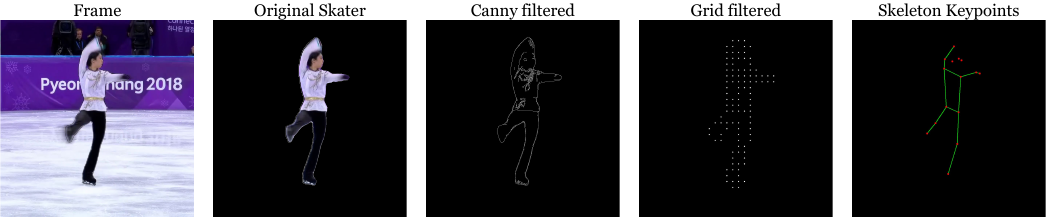}
    \caption{Visual comparison between original frame and various input maps and methods. All images are 1024$\times$1024.}
    \Description{Visual comparison between original frame and various input maps and methods. All images are 1024$\times$1024.}
    \label{fig:qualitative-result}
\end{figure*}

\begin{figure}
    \centering
    \includegraphics[width=1\columnwidth]{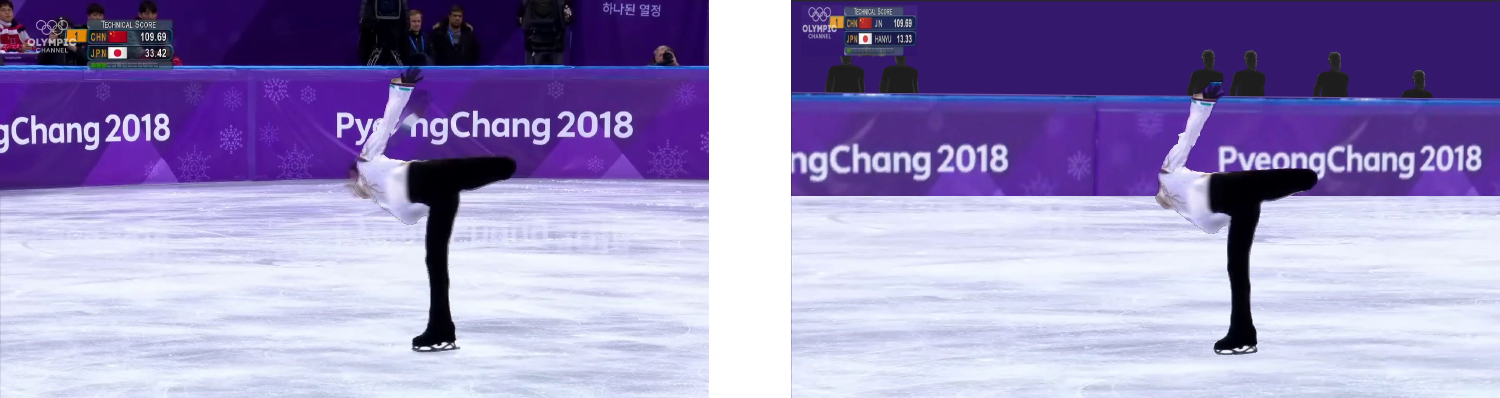}
    \caption{Qualitative comparison between original frame (left) and rendering of the 3D model with applied skater (right).}
    \Description{Qualitative comparison between original frame (left) and rendering of the 3D model with applied skater (right).}
    \label{fig:3d-model}
\end{figure}

\subsubsection*{Content}\label{subsubsec:content}

To validate our approach, we applied the system to a dataset comprising more than \num{40000} frames grouped in \num{48} scenes with duration ranging \qtyrange{9}{55}{\second} extracted from four figure skating performances at \qty{30}{\fps} and 1080p resolution~\cite{olympics2018video}. All videos belong to the 2018 Pyongchang Olympic Games, and thus share key characteristics, including:
\begin{enumerate*}[label=\textit{(\roman*)}]
\item the same ice rink background with predominantly stationary objects and spectators,
\item one figure skater actively moving in the center of the ice rink, and
\item a set of cameras, all following the skater from different points of the ice rink's lower ring.
\end{enumerate*}

\subsubsection*{Setup}

All processing is performed on an Ubuntu 20.04 LTS server equipped with an Intel Xeon Gold 5218 CPU (64 cores, \SI{2.30}{\giga\hertz}) and an NVIDIA Quadro RTX 8000 GPU with \SI{48}{\giga\byte} of memory. The full system is packaged into two Docker~\cite{merkel2014docker} containers: one for the server-side pipeline and another for Unity-based client rendering. The pipeline is implemented in Python using OpenCV, PyTorch, and external tools like COLMAP~\cite{schoenberger2016sfm}, with Unity handling the rendering pipeline on the client side.
Several components of GenStream are fully implemented and tested independently, including actor segmentation, keypoint extraction, inpainting, and data export to 3D scenes. Others, such as camera viewpoint recovery, pose novel challenges and are currently under scrutiny.

\subsubsection*{Actor Segmentation and Frame Decomposition}

Each frame captured by the camera is parsed using YOLOv11~\cite{redmon2016yolo}, which detects all individuals present. Among them, we retain only the largest bounding box to discard bystanders or referees. This bounding box is passed to Segment Anything Model 2 (SAM 2)~\cite{ravi2024sam2}, which produces a fine segmentation mask, as shown in Figure~\ref{fig:qualitative-result} (Original Skater). SAM 2 is a high precision and reliability segmentation model. However, its complexity does not permit real time performance. Alternatively, the detection and segmentation can be performed in one pass by YOLOv11. This approach is much faster, and compatible with real-time live processing, but its output quality is not as robust.
We use the mask to generate, for each frame:
\begin{enumerate*}[label=\textit{(\roman*)}]
  \item actor.png: the actor extracted from the scene, and
  \item background.png: the frame with the actor region masked. 
\end{enumerate*}
The mask is used to add transparency, resulting in RGBA images that clearly differentiate the portions of the frame that belong to the actor and the background, enabling actor-specific compression and background reconstruction.

\subsubsection*{Keypoint extraction}
After detection and segmentation, GenStream proceeds to a pose estimation stage. This step aims to convert each segmented actor into a structured representation of keypoints that compactly captures their motion and posture over time. Since the bounding box of the primary actor has already been computed in the segmentation phase, keypoint extraction is performed in parallel, enabling efficient processing.
To extract keypoints from each actor, we compare three methods with decreasing spatial density: a Canny edge-based filter~\cite{mcilhagga2011canny}, a grid sampling filter, and a YOLOv11 Pose-based skeleton detector~\cite{maji2022yolopose}. The Canny filter highlights contours by detecting high-gradient edges, while the grid filter overlays a regular grid on the edge map and selects representative points from high-detail regions, marking a square’s center when it contains sufficient edge pixels. The YOLOv11 Pose model estimates 17 anatomical landmarks directly and connects them into a human skeleton using predefined edges. All three methods, shown in Figure~\ref{fig:qualitative-result}, are applied to the segmented RGBA image of the actor, ensuring no background interference. Since all approaches yield comparable perceptual quality in downstream tasks, we select the YOLO-based method due to its compact representation and lower bandwidth requirements.

\subsubsection*{Background inpainting}
To enable a clearer view synthesis, image inpainting is used to dynamically fill regions of the background images where actors have been segmented out. Tools such as OpenCV's Telea~\cite{chatterjee2021telea} or Stable Diffusion~\cite{podell2023sdxlstablediffusion} are used depending on available computational resources.
The output is a sequence of image files without foreground actors, to aid the camera viewpoint estimation phase.
While inpainting may introduce some inconsistency across frames, leaving dynamic foreground actors in place typically leads to more severe errors in COLMAP’s reconstruction, as it assumes a static scene. Removing the actors and filling their regions,  even imperfectly,  helps reduce false feature matches and improve the accuracy of the background structure estimation.

\subsubsection*{3D Ice Rink Modeling}
We manually reconstructed an Olympic ice rink in Unity to serve as the scene's geometric and visual reference. This includes accurately placed lines, lighting, and reflection properties to match typical broadcast footage (see Fig.~\ref{fig:3d-model}).

\subsubsection*{Camera Viewpoint Recovery: Challenges and Approaches (Partially Implemented)}
\begin{figure}
    \centering
    \includegraphics[width=1\columnwidth]{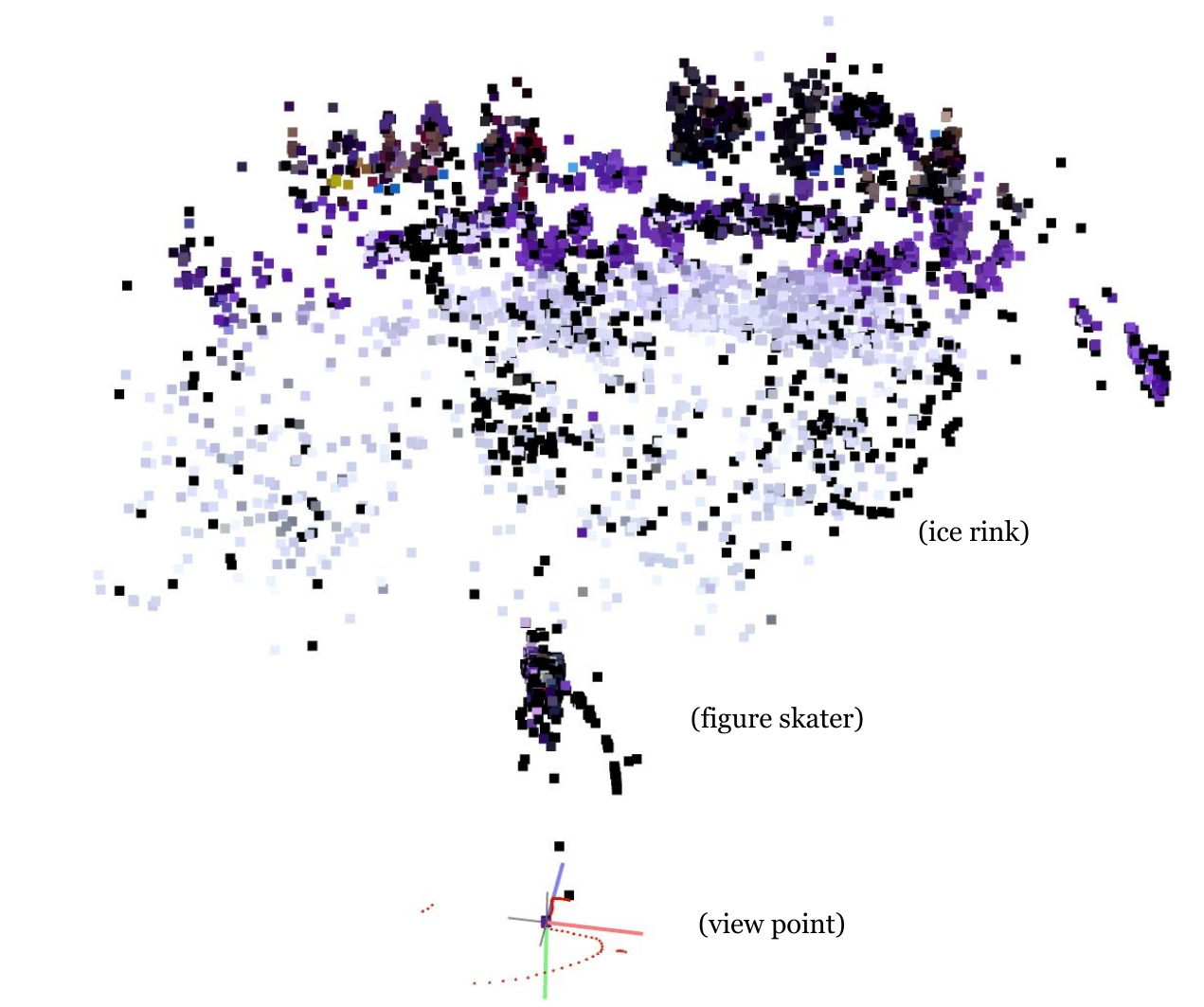}
    \caption{COLMAP generated point cloud of one scene, complete of camera view intrinsics and extrinsics, figure skater in foreground, and ice rink in the background.}
    \Description{COLMAP generated point cloud of one scene, complete of camera view intrinsics and extrinsics, figure skater in foreground, and ice rink in the background.}
    \label{fig:colmap}
\end{figure}

Correctly recovering the camera parameters for each video remains one of the most critical and technically demanding components of GenStream. We explore two alternative approaches:
\begin{enumerate}[label=\textit{(\roman*)}]
    \item \textbf{COLMAP-based 3D Reconstruction (Partially Implemented)} We use COLMAP to generate a sparse point cloud of the static scene and estimate the per-frame camera intrinsics and extrinsics. Figure~\ref{fig:colmap} shows a 2D projection of a reconstruction result. While COLMAP succeeds in generating plausible camera trajectories, integration with Unity remains incomplete. Converting COLMAP's coordinate system and focal length parameters into Unity's camera model introduces geometric inconsistencies that still need to be resolved. Furthermore, to maintain the significant bandwidth reduction that GenStream proposes, COLMAP features extracted from different video scenes need to be fused together into a complete and realistic point cloud model, and sent to the client before the live even takes place. The origin of such video scenes may be the owner of the background stadium, or previous scenes that have been transmitted via regular video streaming.
    \item \textbf{Differentiable Camera Viewpoint Optimization (Partially Implemented)} Another explored approach involves directly optimizing camera parameters by rendering the 3D scene from a randomly initialized camera angle, capturing the resulting view, and comparing it to the inpainted background from the original video frame. The difference between the rendered and inpainted images serves as a loss function, which is minimized through iterative optimization to estimate the correct camera viewpoint. Efforts in this direction have raised many challenges, both in terms of implementation structure and resulting quality. Furthermore, the computation required for such an optimization is not compatible with live streaming. To address this last challenge, we implemented a more efficient solution: we perform the full optimization only for the first frame of each scene, then use a SIFT-based relative pose estimation~\cite{lowe2004sift} to propagate camera viewpoints to subsequent frames by comparing each one to its immediate predecessor. This significantly reduces the overall computational burden while maintaining adequate tracking accuracy.
\end{enumerate}

\subsubsection*{Data export (Partially Implemented}
Finally, the extracted data, comprising actor bounding boxes and keypoints, and camera viewpoints (or optionally COLMAP's complete scene information), is saved to a CSV file and then encoded for transmission.

With regards to keypoints transmission, each coordinate is encoded as a 11-bit unsigned integer, sufficient to represent pixel positions within a 1080p resolution frame (up to 1920 pixels). 
Crucially, because the number and ordering of keypoints are fixed and known in advance, each segment of the bitstream corresponds to a specific coordinate of a specific object in a deterministic way. 
For example, the first 11\,bits always encode the X-coordinate of the first keypoint, the next 11\,bits the corresponding Y-coordinate, and so on. 
This layout removes the need for additional metadata, enabling both tight compression and extremely fast parsing on the client side. 
If an actor is not detected in a particular frame, a sentinel control value (2000, greater than 1920 and thus outside the possible coordinate range) is inserted in place of the missing coordinates. 
This allows the client to gracefully handle missed detections while maintaining alignment with the expected bitstream format.
However, to develop such a tight compression scheme for COLMAP's features, it is required to solve the related challenges mentioned previously.
It is worth noting that the compressed data is delivered to the client at intervals, aligned with the target latency requirements of live video streaming applications.

\subsubsection*{Generative Model Training Setup}
To generate realistic figure skater images from skeleton keypoints, we train a cGAN which receives in input paired image data, where each sample consists of two components: a reconstructed image of the person based on their keypoints, and the corresponding ground truth image. These two images, originally sized at either \vxv{128}{128}, \vxv{256}{256}, or \vxv{1024}{1024}, are concatenated to form paired input images of size \vxv{256}{128}, \vxv{512}{256}, and \vxv{2048}{1024}, respectively.
The input images, split into \qty{80}{\percent} for training and \qty{20}{\percent} for testing, are augmented using random resizing, cropping, and mirroring to enhance robustness.

\subsubsection*{Client-Side Reconstruction}
GenStream’s client-side rendering is implemented in Unity and currently uses the manually constructed 3D model of the Olympic ice rink shown in Figure~\ref{fig:3d-model} as its background rendering approach. To complete the synthetic view, the generated figure skater is placed at the correct spatial location along the camera’s optical axis, ensuring that their size and position match the original video.
In our current implementation, camera parameters are inferred manually to approximate the original view. Using these values, we can overlay the generated figure skater and achieve a visually plausible frame composition on top of the 3D background. However, this manual solution is clearly not scalable, and useful only for qualitative presentations. Once this challenge is solved, the Unity renderer is already set up to accept per-frame camera values.

\section{Conclusion}

GenStream introduces a framework for video streaming, one that replaces dense pixel transmission with high-level semantic instructions that guide client-side generative reconstruction. By transmitting only sparse skeleton-based keypoints and camera viewpoint metadata, GenStream achieves a bandwidth reduction of over 99.9\% compared to conventional HEVC encoding, while preserving essential perceptual cues for immersive viewing.
This system provides a blueprint for a new generation of streaming architectures that prioritize semantic content over pixel fidelity. However, several open challenges remain before GenStream can reach its full potential.

As highlighted by our analysis, the most immediate challenge is the significant computational load shifted to the client. Making GenStream practical requires substantial research into lightweight, efficient generative models suitable for resource-constrained devices. This is a primary axis for future work.

A key future direction is the evolution from 2D sprite-like renderings to fully volumetric, viewpoint-consistent representations. Instead of compositing a 2D actor into a 3D scene, we envision modeling the actor as a moving 3D skeleton driving a volumetric avatar. This would allow the system to support free-viewpoint video synthesis, enabling clients to explore scenes from arbitrary perspectives, even those not originally captured by any camera.
Realizing this vision introduces two critical challenges. First, it requires the fusion of multiple partial point clouds captured from different video sequences into a canonical 3D representation of the actor and background. Second, it demands lifting 2D skeletons to reliable 3D joint configurations in single-camera scenarios. These challenges intersect fields such as multi-scene registration, 3D human reconstruction, and sparse generative modeling.
Furthermore, while our current implementation uses Unity for client-side rendering, future deployments might benefit from alternative engines or custom viewers that better align with emerging hardware and web standards. Importantly, GenStream remains renderer-agnostic; any tool capable of interpreting the structured metadata and synthesizing scenes from it could serve as the playback environment.
Another important avenue lies in extending GenStream beyond the controlled setting of figure skating to more complex, multi-agent scenes such as football or gymnastics. These scenarios involve richer interactions, dynamic camera work, and frequent occlusions, each posing new demands on segmentation, multi-actor pose estimation, handling occlusions, and synthesizing interactions between generative avatars. Robustly scaling to such complexity is a non-trivial research problem.
Finally, we note that background modeling using COLMAP or similar SfM tools requires careful fusion of features from multiple scenes to build a unified point cloud representation. These features can originate from the event organizer’s asset library or be extracted from previously streamed footage. Solving this fusion problem efficiently and compactly is crucial for pre-distributing 3D venue models to clients.

GenStream is a first step toward a future where semantic understanding and generative capabilities replace brute-force video transmission. By offloading reconstruction to powerful client-side models and transmitting only what truly matters, it offers a scalable, perceptually meaningful approach to media delivery for the post-codec era.

\section{Acknowledgments}
\balance
The financial support of the Austrian Federal Ministry for Digital and Economic Affairs, the National Foundation for Research, Technology and Development, and the Christian Doppler Research Association, is gratefully acknowledged. Christian Doppler Laboratory ATHENA: https://athena.itec.aau.at/

\bibliographystyle{ACM-Reference-Format}
\bibliography{ref}





\end{document}